\documentclass[conference]{IEEEtran}
\IEEEoverridecommandlockouts

\usepackage{cite}
\usepackage{amsmath,amssymb,amsfonts}
\usepackage{algorithmic}
\usepackage{graphicx}
\usepackage{textcomp}
\usepackage{xcolor}
\usepackage{comment}
\usepackage{subcaption}
\def\BibTeX{{\rm B\kern-.05em{\sc i\kern-.025em b}\kern-.08em
    T\kern-.1667em\lower.7ex\hbox{E}\kern-.125emX}}
\begin{document}

\title{Performance Monitoring of Proton Exchange Membrane Water Electrolyzer by Transformers-Based Machine Learning Model
}

\author{Bingqing Chen$^{*}$ \quad Ivan Batalov$^{*}$ \quad Qiu Chen \quad Weiqi Ji \quad Lei Cheng
\thanks{*These authors contributed equally. }
\thanks{This work was performed during when the authors are/were employed by Bosch Research \& Technology Center.} 
\thanks{Contact: {\tt\footnotesize bingqing.chen@us.bosch.com}
}%
\vspace{-0.2cm}
}

\maketitle

\begin{abstract}
 Green hydrogen plays an essential role in decarbonization, with capacity projected to scale to 560GW by 2030 (vs. 1.39GW in 2023) in net‑zero settings. Proton exchange membrane (PEM) electrolysis is one of the most promising technology routes to green hydrogen production, and real-time system health monitoring of PEM electrolyzers is essential for their scalable
deployment. In lab settings, performance degradation can be characterized through electrochemical
testing protocols by periodic pauses of normal operation. Such interruption is not practical for full-scale stack deployments, limiting system operators' ability to make real‑time assessment of state‑of‑health (SoH). We present a machine learning (ML) framework that performs virtual electrochemical characterization during normal operation. The method uses an encoder–decoder transformer, conditioned on operational data to reconstruct characterization outputs, focusing here on polarization curves. Inspired by patch‑based sequence tokenization, we segment the inputs into patches and encodes them to form meaningful tokens, which substantially improves learning efficiency. Across four longitudinal runs, lasting up to 478 hours on different test cells and loading cycles, the model accurately reconstructed polarization curves and achieved 10$\times$ reduction in mean squared error (MSE) compared to a vanilla transformer. This proof‑of‑concept demonstrates that ML models can enable continuous performance monitoring for PEM electrolyzers and that the encoder captures meaningful latent representation of SoH, opening up opportunities to derive interpretable  indicators in future work.
\end{abstract}

\maketitle

\section{Introduction}
Hydrogen, a high-energy-density carrier,
 is increasingly recognized as an essential pathway for decarbonization, particularly in sectors where direct electrification is challenging, such as heavy industry, long-haul transportation, and high-temperature industrial processes \cite{satyapal2023us}. Furthermore, hydrogen can support grid balancing in the built environment, by complementing electrified heating systems (e.g., hybrid heat pump) during peak demand \cite{cardenas2025heat,kurniawati2025integrated}. 
 \textit{Green hydrogen} is produced via water electrolysis powered by renewable electricity.  A total of 560 GW capacity by 2030 (1.39 GW capacity in 2023) is required in Net Zero Emissions by 2050 (NZE) scenario \cite{iea}. Proton exchange membrane (PEM) electrolysis is among the most promising technology routes to  green hydrogen production, due to its ability to operate efficiently at high current density and produce high-purity hydrogen \cite{salehmin2022high, sezer2025comprehensive}. Furthermore, it has a fast ramp rate and a wide partial-load operating range, making it ideal for coupling with variable renewable energy sources, such as wind and solar \cite{crespi2023experimental}. 

\begin{figure*}
    \centering
    \includegraphics[width=0.9\linewidth]{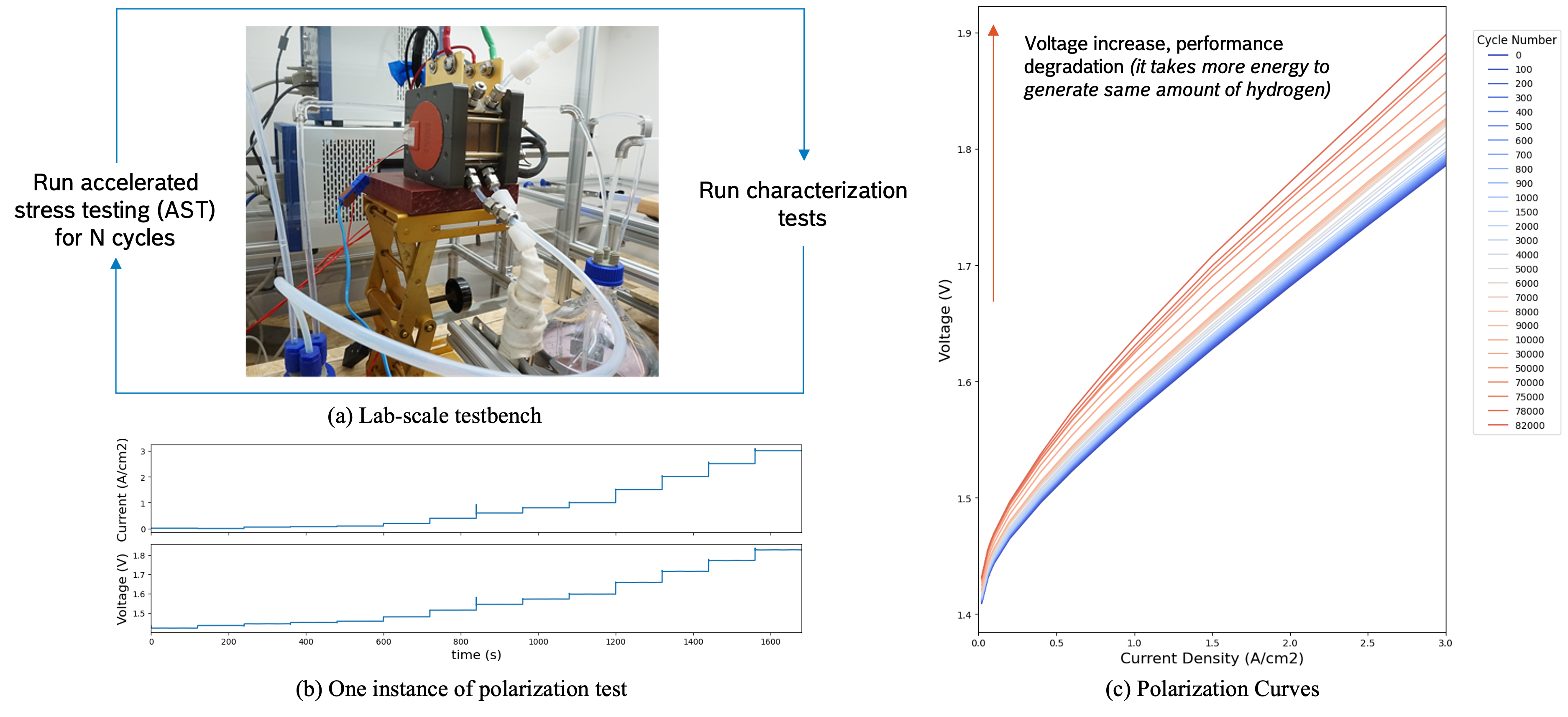}
    \caption{Testbench and performance characterization via polarization curves.}
    \label{fig:lab}

\end{figure*}

Among challenges for cost-effective and scalable green hydrogen production, real-time system health monitoring is essential for the practical deployment of PEM electrolyzers \cite{woelke2025predicting}. In lab-scale devices, performance degradation can be assessed through controlled, pre-defined electrochemical testing protocols by periodic pauses from normal operation, such as polarization curve measurement and electrochemical impedance spectroscopy (EIS). However, full-scale stack deployments (in MW) often operate under dynamic conditions due to their coupling with renewable electricity, where such controlled pauses may not be practical. This limits system operators' ability to access the system's state-of-health (SoH) in real-time operation. Despite incomplete physical understanding of the aging mechanisms, ML can bridge this gap by learning from data. Such data-driven models can function as surrogates for physical systems, enabling virtual testing and continuous performance monitoring.
There is a small but growing body of works that use AI to model PEM electrolyzer \cite{ozdemir2024performance, hayatzadeh2024machine, chen2024machine}. While these works show promising results under static operating conditions using conventional machine learning techniques (such as support vector machine \cite{ozdemir2024performance} and shallow neural network \cite{hayatzadeh2024machine}), these models may not have the capacity to learn realistic, dynamic loading cycles, e.g., wind and solar profiles,  underscoring the need for more advanced approaches (e.g., transformer-based models) to robustly monitor PEM electrolyzer performance.

In this work, we formulate the problem of performance monitoring as a sequence-to-sequence task, and leverage an encoder-decoder transformer architecture \cite{vaswani2017attention} for flexible conditional generation. 
Specifically, our model is trained to reconstruct electrochemical characterization tests, i.e., polarization curves, conditioned on operational data. Inspired by PatchTST \cite{nie2023timeseriesworth64}, we segment the time series data into overlapping patches. Then we project each patch, containing multivariate information, into a single token to encode current-voltage relationship of the electrochemical systems.  These techniques lead to a 10-fold reduction in mean squared error (MSE), compared to a vanilla transformer \cite{vaswani2017attention} model. We evaluated our method on four longitudinal experiments, lasting up to 478 hours, and demonstrated its ability to accurately reconstruct polarization curves; Such technology would allow system operators to monitor real-time performance without interrupting normal operation, and reduce the costs of running diagnostic cycles.


\section{Testbench and Experiment Protocols}\label{sec:exp_setup}
Electrochemical testing and measurements were carried out in a home-built testbench  with  a re-circulating water loop with ion-exchanging resin filter,  as shown in Figure \ref{fig:lab}(a). Water inlet and cell temperature are both controlled at 80\textsuperscript{o}C.  The membrane electrode assembly was assembled in a cell hardware using Pt- and Au-coated Ti flow field on the anode and cathode sides, respectively. The catalyst-coated membrane was laminated down to a 5 cm\textsuperscript{2} active area. Ti fiber based porous transport layer (PTL) with Pt coating were used on anode side. Carbon fiber based gas diffusion layer (GDL) without micro porous layer was used on cathode side. A combination of polytetrafluoroethylene (PTFE) gaskets were used on cathode and anode side to ensure approximately 20\% thickness compression on the GDL side and no thickness compression on PTL side.  The electrochemical testing was conducted using a Gamry Ref 3000 potentiostat with a 30A current booster. 
During electrochemical testing, temperature, pressure, and flow rate are maintained at constant values, whereas either the current or the voltage is controlled.  Accelerated stress tests (ASTs) by voltage cycling was used to induce performance degradation. Standard electrochemical charaterization measurements, including polarization curves, were collected at irregular intervals of ASTs cycles; characterization tests were conducted more frequently at beginning of life, when more degradation was expected.

\begin{figure*}[t!]
\centering
\begin{subfigure}{.53\textwidth}
  \centering
  \includegraphics[width=\linewidth]{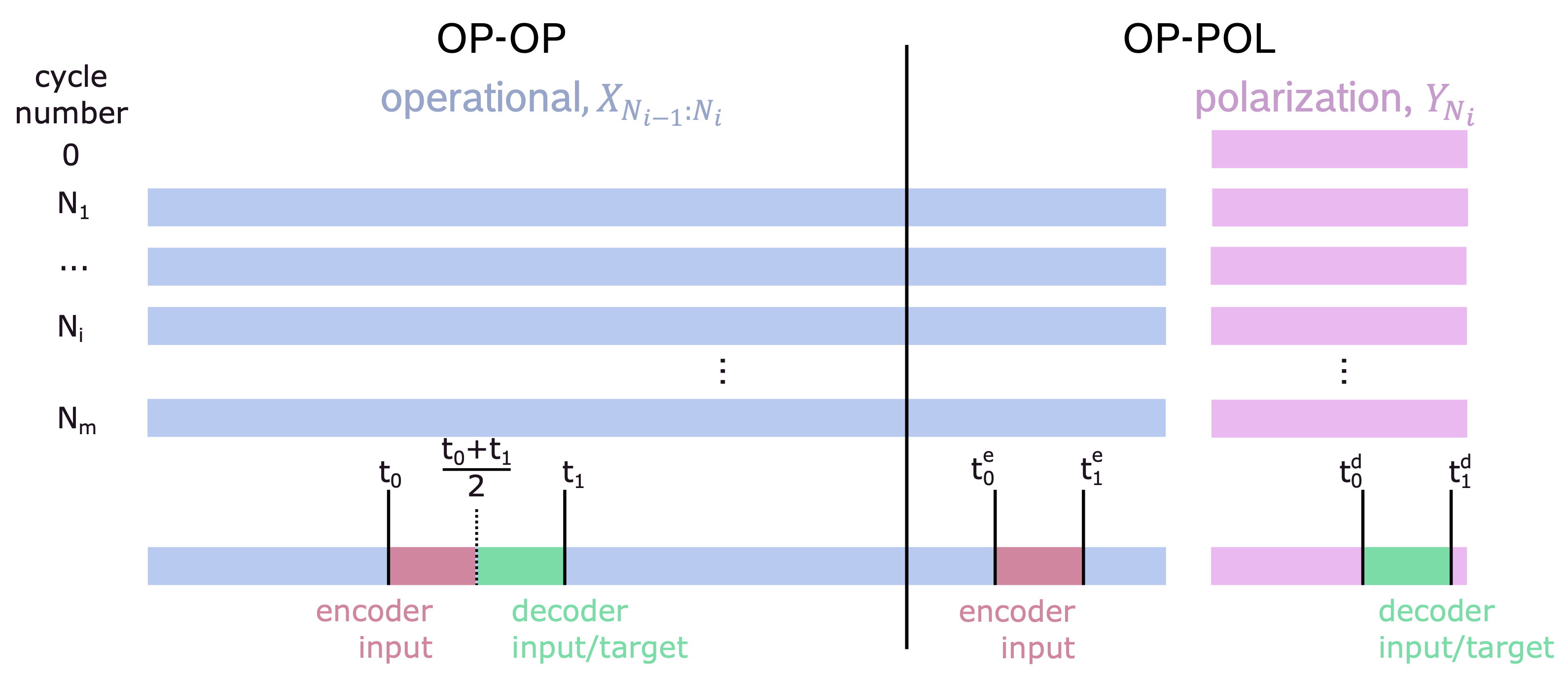}
  \caption{Description of Dataset}
  \label{fig:data}
\end{subfigure}%
\begin{subfigure}{.47\textwidth}
  \centering
  \includegraphics[width=\linewidth]{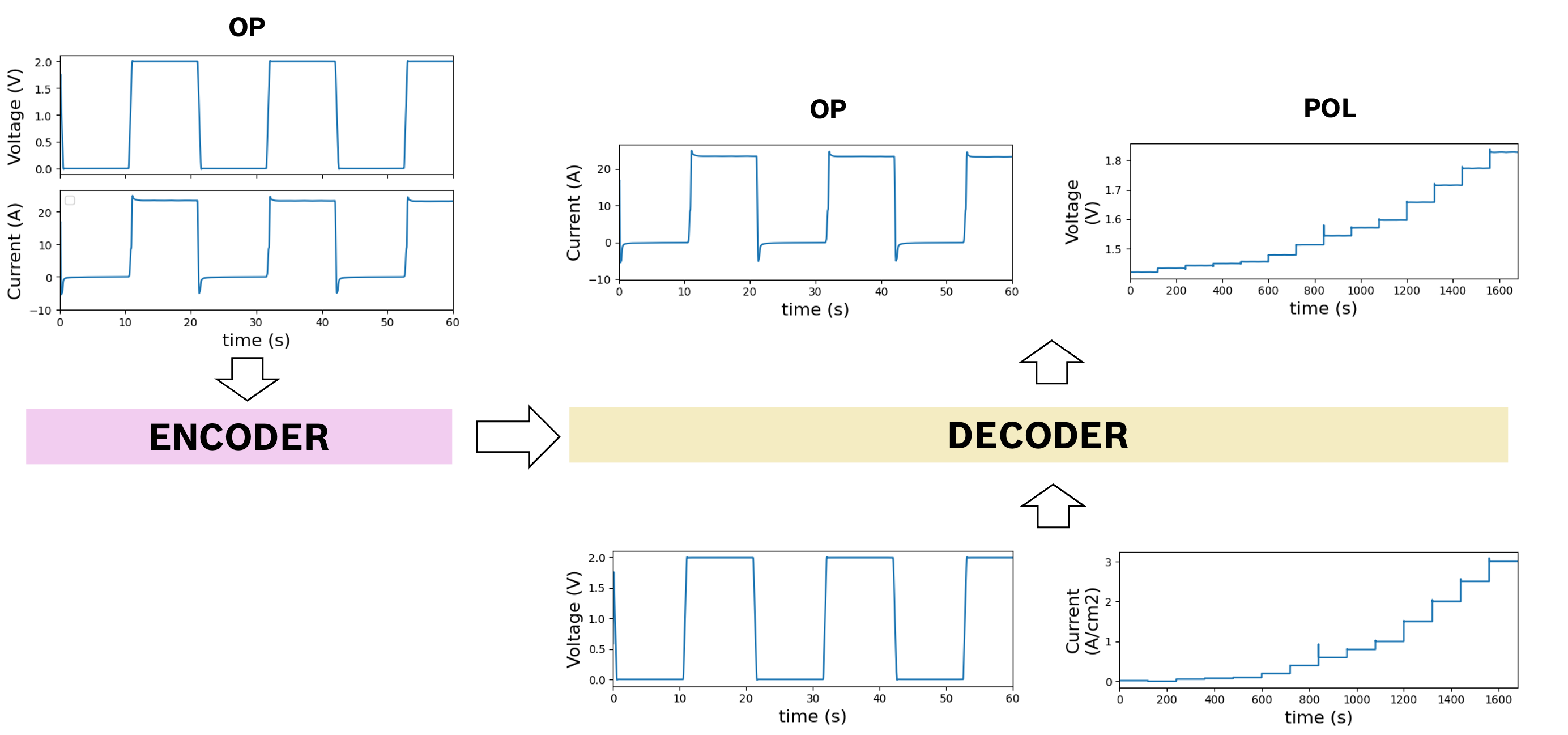}
  \caption{Example Data}
  \label{fig:problem}
\end{subfigure}
\caption{Problem Setup. (a) We preprocess the data into paired samples of operational data (OP), $X_{N_{i-1}:N_i}$, and polarization test (POL), $Y_{N_i}$. Shaded regions indicate the encoder input (pink) and the decoder input/target (green). 
(b) The  decoder predicts the target channel of either polarization curve (OP–POL) or operational data (OP–OP) given the input channel, conditioned on operational data received by the encoder.}
\label{fig:problem_formulation}
\end{figure*}

A polarization curve characterizes the current-voltage relationship of a electrochemical cell. 
In electrochemical processes, electrochemical potential\footnote{Cell voltage is the electrochemical potential difference between anode and cathode.} is an expression of electron energy that drives chemical reactions, and the geometric current density\footnote{Geometric current density is current normalized by electrode active area.} is an expression of related reaction rate \cite{bard2022electrochemical}.  A polarization test involves subjecting the cell to a series of step-wise controlled geometric current density levels, current density maintained for 2 minutes for each level, and recording the corresponding steady-state voltage values, as exemplified in Figure \ref{fig:lab}(b). One can use polarization curves to characterize performance degradation, by pausing normal operation of the PEM electrolysis cell or ASTs at different time stamp during the device lifetime. In Figure \ref{fig:lab}(c) as an example, typically at the same current density, the cell voltage increases with the number of AST cycles accumulates. This is indicative that for the same geometric current densities generated to produce hydrogen gas, a higher cell voltage (and energy) is required to drive the electrochemical reaction as a result of performance degradation.

\section{Methodology}
\label{sec:method}

We formulate learning the current-voltage relationship of electrochemical cells as a time-series modeling problem. We learn a transformer-based model that takes a snapshot of operational time series as inputs and predicts the polarization tests. We first formulate the problem setup and then present the model architecture.

\subsection{Problem Setup} \label{sec:dataset}

First, we need to preprocess the dataset to be amenable to ML. 
Recall that polarization tests are conducted at irregular intervals; Let $\mathcal{N} = \{0, N_1, \dots, N_m\}$ be the set of loading cycles after which characterization tests are performed. 
The data at the $i^\text{th}$ loading cycle is denoted as $X_i \in \mathbb{R}^{L \times D}$, where $L$ is the cycle length and $D$ the measurement dimension. We denote the concatenation of data between cycles $N_{i-1}+1$ and $N_i$  as $X_{N_{i-1}+1:N_i} = \text{concat}(X_{N_{i-1}+1}, \dots, X_{N_i})\in \mathbb{R}^{(N_i-N_{i-1})L \times D}$, and the polarization test data at the $N_i^\text{th}$ cycle as $Y_{N_i} \in \mathbb{R}^{T \times D}$. 
After preprocessing, there is paired relationship between the operational data (OP), $X_{N_{i-1}+1:N_i}$ and the polarization test data (POL), $Y_{N_i}$. While the cells went through AST cycles designed to induce degradation, we observe the method to be broadly applicable to other loading cycles, such as wind and solar profiles, in later experiments to be covered in a follow-up work. In this work, the measurements consist of voltage and current density measured at 10Hz; therefore $D=2$ since other variables are fixed. The ML framework, however, naturally extends to higher-dimensional data.

As illustrated in Figure \ref{fig:data}, we use two data mixtures for training. The primary objective is to predict voltage, given controlled current during polarization test (OP-POL). We also train the model to predict current given controlled voltage during operation (OP-OP) as an auxiliary objective. The decoder takes the controlled channel as input, and predicts the target channel of either operational or polarization test, as illustrated in \ref{fig:problem}. To highlight an important detail, the ASTs are voltage-controlled, while the polarization tests are current-controlled. So different channels are used as input and output for OP-POL and OP-OP. 

\subsection{Model Architecture}

\begin{figure}
    \centering
    \includegraphics[width=\linewidth]{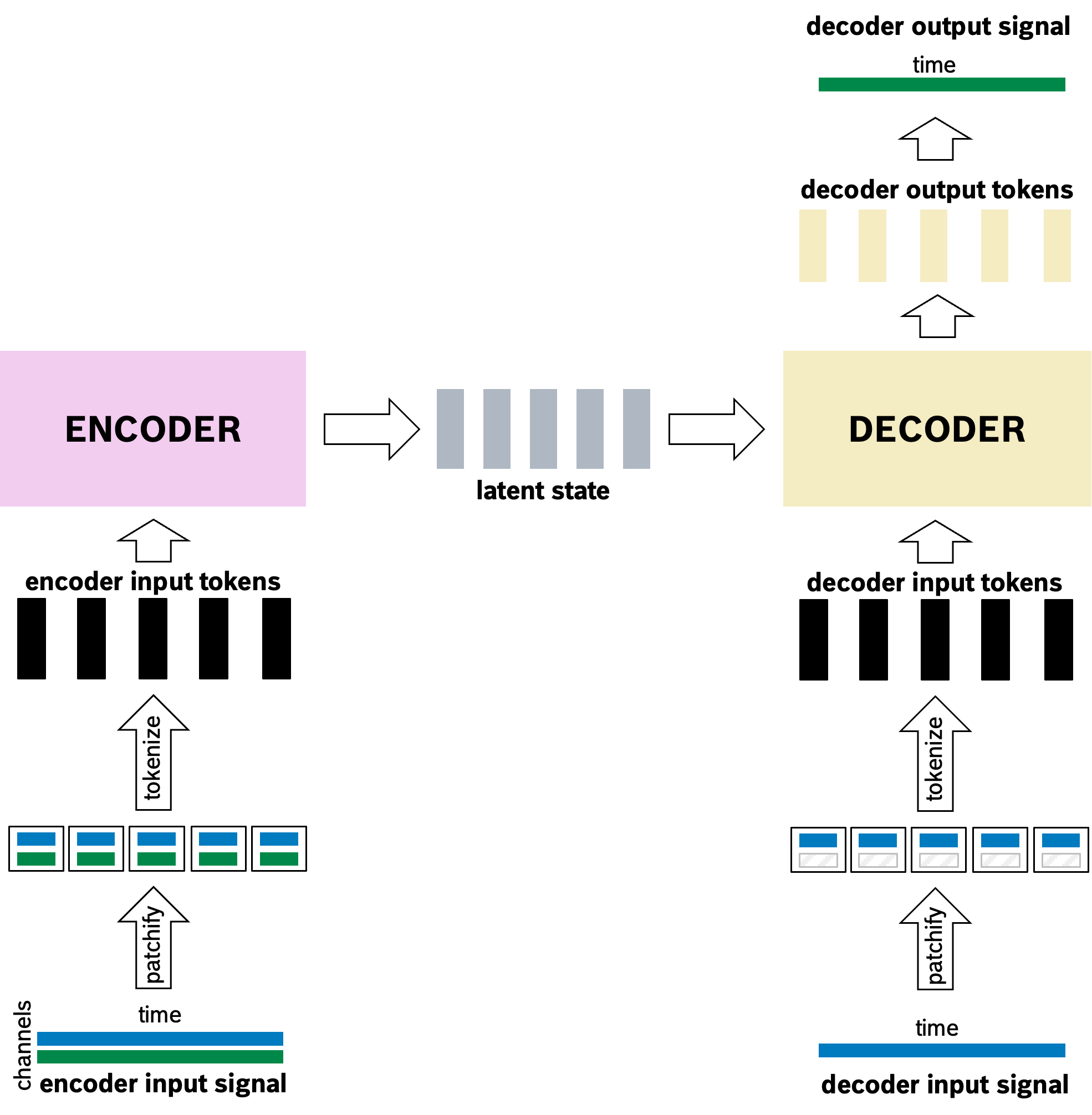}
    \caption{Patch Transformer}
    \label{fig:patch_transformer}
\end{figure}

Instead of the more popular paradigm of the decoder-only transformer, trained with an autoregressive objective \cite{das2024decoder}, we adopt an encoder-decoder transformer with a reconstruction objective (Figure \ref{fig:patch_transformer}) to allow for flexible conditional generation, i.e., predicting polarization test conditioned on operational data (OP-POL), along with the auxiliary task of reconstructing operational data (OP-OP). 

The model consists of a transformer-based encoder  $f_{\text{enc}}$, a trans-former-based decoder $f_{\text{dec}}$, and a patch-based sequence tokenizer (Figure \ref{fig:patch_transformer}). To recap, data pairs in OP-OP and OP-POL are illustrated in Figure \ref{fig:data} and how they relate to the model is shown in Figure \ref{fig:problem}. Formally, we denote the dataset as $\mathcal{D} = \mathcal{D}_{\text{OP-OP}} \cup \mathcal{D}_{\text{OP-POL}}$. The encoder $f_{\text{enc}}$ takes an input sequence $X_{\text{enc}}$ and produces a latent representation, $Z = f_{\text{enc}}(X_{\text{enc}})$. The decoder $f_{\text{dec}}$ receives both the latent embedding $Z$ and a input sequence $X_{\text{dec}}$ to predict the target channel $\hat{Y}_{\text{dec}} = f_{\text{dec}}(X_{\text{dec}}; Z)$. The learning objective is the MSE loss (Eqn. \ref{eq:objective}) between the prediction and target sequence, where $\theta$ represents learnable parameters in the transformer. 
\begin{align}\label{eq:objective}
\mathcal{L}(\theta; \mathcal{D}) = \mathbb{E}_{X_{\text{enc}}, X_{\text{dec}}, Y\sim \mathcal{D}}=||\hat{Y}_{\text{dec}}-Y||_2^2
\end{align}

\textit{Patching.} Inspired by the successful PatchTST architecture \cite{nie2023timeseriesworth64}, we segment the long time series to subseries-level patches as illustrated in Figure \ref{fig:patch_transformer}. This patching design compresses the raw sequence into a shorter sequence of patches and preserves local semantic information. We denote the patch length as $p$ and the stride as $s$. For a sequence length $l$, the number of patches is 
$N_{\text{patches}} = \left\lfloor \frac{l - p}{s} \right\rfloor + 1$.
In our implementation, we use $l=1024$, $p=64$, and $s=32$, and thus $N_{\text{patches}} = \left\lfloor \frac{1024 - 64}{32} \right\rfloor + 1 = 31$. With the use of patches, the number of input tokens is reduced by roughly $s$ times and thereby reduces the computational complexity of attention mechanism quadratically by $s^2$. Despite the reduced computation cost, this design choice improves performance in our experiments, as it enables the transformer modules to capture long-range context. 

\textit{Channel Mixing.} In contrast to the channel-independence strategy in PatchTST,  we mix the channels to learn the current-voltage relationship in OP-OP and OP-POL. After we segment the sequence into patches, we project each patch into a token, mixing the information between channels. Since the experiments can be either current- or voltage-controlled, it varies which channel is used as input or output. We keep consistent ordering of the channels and pad the unavailable channel, also illustrated in Figure \ref{fig:patch_transformer}.


\textit{Implementation Details.} We used the same set of hyperparameters for all the experiments and got consistent performance without hyperparameter tuning. We use the Adam optimizer \cite{kingma2014adam} with a learning rate of $1\times 10^{-4}$ and reduce the learning rate on plateau. We use a batch size of 32 and train for 100 epochs. For train-validation split, we place every third characterization test cycle to the validation set, i.e., data corresponding to $\{N_1, N_2, N_4, N_5, \cdots\}$ go to the training set, and those corresponding to $\{N_3, N_6, \cdots\}$ go to the validation set. All results are reported on the validation set. Since the decoder predicts a fixed sequence length of $l=1024$, we predict a polarization test over as a moving window, concatenate the predictions, and then aggregate the predictions into a polarization curve with the same protocol as physical measurements.

\section{Results}
\label{sec:results}
We evaluate the proposed method on four experiment runs following the experiment protocol described in Section \ref{sec:exp_setup}. Models are trained independently for each experiment run. The four experiment runs  last for 478 hours. Run 1 is operated over on/off cycles (0-2 V) and Run 2-4 are operated over loading/unloading cycles (1.45-2 V). The prediction error as measured by mean squared error is shown in Table \ref{tab:summary}. While the two loading cycles represent distinct aging mechanisms, transformer-based models are able to perform consistently well on all of them. Using the patching technique with the transformer significantly improves performance, especially on polarization curve prediction, where the prediction error decreases by more than one order of magnitude on all four runs. In comparison, the vanilla transformer makes noisy predictions, and thus we believe the improvement arises from patch transformer's capability to retain local semantic information \cite{nie2023timeseriesworth64}.

\begin{figure}[h]
    \centering
    \includegraphics[width=\linewidth]{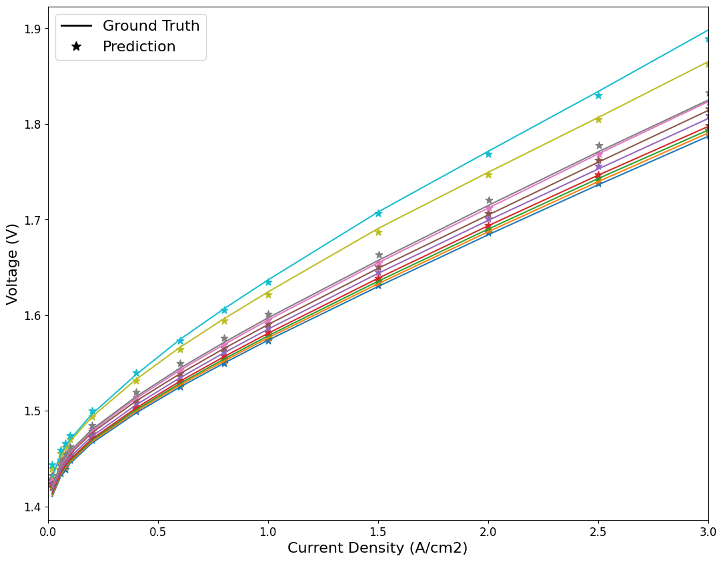}
{%
  \caption{Ground Truth vs. Predicted Polarization Curves (Run 1; Validation Set)}%
  \label{fig:pol_compare}
}
\end{figure}

Figure \ref{fig:pol_compare} shows the qualitative comparison of ground truth vs. predicted polarization curves by patch transformer. It shows the promising result that patch transformer can accurately predict polarization curves. This opens up the possibility for system operators to perform characterization tests virtually without interrupting normal operation. 

In our problem setup, the model only sees a snapshot of current operational data and has no access to historical context, including how long the system has been running since beginning of life. The transformer model is able to spot subtle different in current-voltage relationship due to degradation and accurately predict polarization curves. 
It is also worth noting that the decoder input is the same step-wise current density profile controlled experimentally. Thus, the information on performance degradation must come from the latent state $Z$. This implies that the encoder is able to extract meaningful  state-of-health information from operational data.

\begin{table}[]
  \caption{Summary of Prediction Error (Validation Set)}
\centering
   \begin{tabular}[t]{lcccc}
\hline
\multicolumn{5}{c}{\textbf{AST curve prediction error}} \\
\hline
 & Run 1 & Run 2 & Run 3 & Run 4 \\
\hline
Transformer \cite{vaswani2017attention} & 0.006 & 0.016 & 0.015 & 0.028 \\
Ours  & \textbf{0.005} & \textbf{0.010} & \textbf{0.002} & \textbf{0.021} \\
\hline
\multicolumn{5}{c}{\textbf{Polarization curve prediction error}} \\
\hline
 & Run 1 & Run 2 & Run 3 & Run 4 \\
\hline
Transformer \cite{vaswani2017attention} & 0.005 & 0.027 & 0.006 & 0.014 \\
Ours   & \textbf{0.0001} & \textbf{0.001} & \textbf{0.0001} & \textbf{0.001} \\
\hline
\end{tabular}
{%
  \label{tab:summary}
}
\end{table}

\section{Conclusions}
We developed a proof-of-concept to show that ML models can enable real-time performance monitoring of PEM electrolyzer. We can conduct polarization tests virtually with ML models conditioned on operational data. We demonstrated that transformer-based models can work consistently well across different test cells and loading cycles. Furthermore, we improved on the original transformer \cite{vaswani2017attention} by incorporating the patching technique \cite{nie2023timeseriesworth64} and encoding of current-voltage relationship, which reduced prediction error by more than a magnitude. 

As future work, we will train our models on more data from test cells or stacks as they become available. We will also evaluate generation capabilities over time and over more diverse loading cycles. It is promising that the encoder-decoder transformer is able to characterize system performance with only a snapshot of current operational data and allows us to extract meaningful state-of-health information. We will explore how to extract low-dimensional, interpretable indicators that system operators can use from high-dimensional latent state.

\bibliographystyle{IEEEtran}
\bibliography{reference}

\end{document}